\documentclass[10pt,twocolumn,letterpaper]{article}

\usepackage{cvpr}              % To produce the CAMERA-READY version
\usepackage{bm}

\definecolor{cvprblue}{rgb}{0.21,0.49,0.74}
\usepackage[pagebackref,breaklinks,colorlinks,allcolors=cvprblue]{hyperref}
\usepackage{cuted}
\usepackage{enumitem}
\usepackage{float}
\usepackage{tikz}
\usepackage{booktabs,multirow}
\usepackage{dblfloatfix}
\usepackage[dvipsnames]{xcolor}
\definecolor{deepblue}{RGB}{50,161,219}
\definecolor{deepred}{RGB}{237,28,36}
\definecolor{pumpkin}{HTML}{FC7A1E}
\definecolor{cornflower}{HTML}{758BFD}

\def\paperID{} % *** Enter the Paper ID here
\def\confName{CVPR}
\def\confYear{2026}

\title{Next-Scale Autoregressive Models for Text-to-Motion Generation}

\author{
Zhiwei Zheng\quad
Shibo Jin \quad Lingjie Liu \quad Mingmin Zhao \\
University of Pennsylvania \\
}

\usepackage[most]{tcolorbox}
\tcbuselibrary{skins,breakable}
\newtcolorbox{moduleprompt}[1][]{
  enhanced,
  breakable,
  colback=gray!5,
  colframe=gray!50,
  boxrule=1pt,
  arc=2pt,
  boxsep=2pt,
  left=4pt, right=4pt, top=4pt, bottom=4pt,
  width=\linewidth,
  #1% allow overrides
}

\begin{document}

\maketitle

% \begin{strip}
% \vspace{-30pt}
% \centering
% \includegraphics[width=\textwidth]{figure/teaser-latest.pdf}
% \vspace{-20pt}
% \captionof{figure}{MoScale accurately captures global semantic structure in text descriptions—such as \textcolor{deepred}{\emph{two}} \textcolor{deepblue}{\emph{jumping jacks}} and \textcolor{deepred}{backward} \textcolor{deepblue}{zig-zag}—where prior diffusion-, masked-transformer–, and next-token autoregressive models fail. Its next-scale autoregressive design with hierarchical causality enables MoScale to preserve these long-range semantics while maintaining realistic motion.}
% \label{fig:teaser}
% \vspace{-5pt}
% \end{strip}

\begin{strip}
\vspace{-30pt}
\centering
\begin{tikzpicture}
\node[anchor=south west, inner sep=0] (teaser-figure) at (0,0)
{\includegraphics[width=\textwidth]{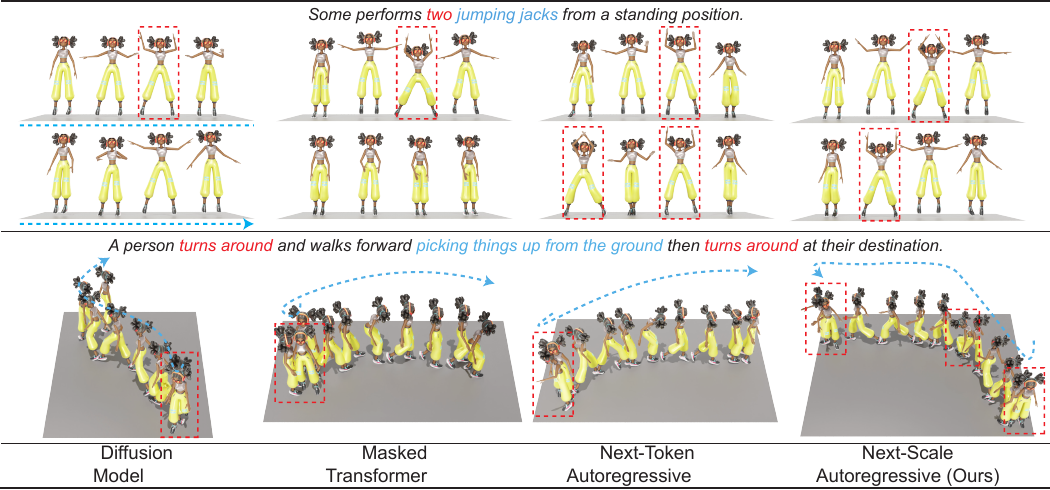}};
\begin{scope}[x={(teaser-figure.south east)}, y={(teaser-figure.north west)}]
\node[anchor=west, text=black, font=\small, inner sep=0pt, outer sep=0pt]
at (0.139, 0.029)
{~\cite{zhang2023remodiffuse}};
\node[anchor=west, text=black, font=\small, inner sep=0pt, outer sep=0pt]
at (0.406, 0.029)
{~\cite{guo2025snapmogen}};
\node[anchor=west, text=black, font=\small, inner sep=0pt, outer sep=0pt]
at (0.658, 0.029)
{~\cite{zou2024parco}};
\end{scope}
\end{tikzpicture}
\vspace{-20pt}
\captionof{figure}{MoScale accurately captures global semantic structure in text descriptions, such as \textcolor{deepred}{\emph{two}} \textcolor{deepblue}{\emph{jumping jacks}} and sequential actions including \textcolor{deepred}{turn around}, \textcolor{deepblue}{pick things up}, and \textcolor{deepred}{turn around}, where prior methods fail to align with the text. Our next-scale autoregressive design with hierarchical causality enables MoScale to preserve these long-range semantics while maintaining realistic motion.}
\label{fig:teaser}
\vspace{-5pt}
\end{strip}

% MoScale
\newcommand{\moscale}{\textit{MoScale}\xspace}
\newcommand{\numsup}[2]{\ensuremath{#1^{\scriptsize #2}}}

\begin{abstract}
Autoregressive (AR) models offer stable and efficient training, but standard next-token prediction is not well aligned with the temporal structure required for text-conditioned motion generation. We introduce MoScale, a next-scale AR framework that generates motion hierarchically from coarse to fine temporal resolutions. By providing global semantics at the coarsest scale and refining them progressively, MoScale establishes a causal hierarchy better suited for long-range motion structure. To improve robustness under limited text–motion data, we further incorporate cross-scale hierarchical refinement for improving per-scale initial predictions and in-scale temporal refinement for selective bidirectional re-prediction. MoScale achieves SOTA text-to-motion performance with high training efficiency, scales effectively with model size, and generalizes zero-shot to diverse motion generation and editing tasks. Code and additional materials are available on the \href{https://zhiwei-zzz.github.io/MoScale/}{webpage}. 
\end{abstract}
\section{Introduction}
\label{sec:intro}
Text-to-motion generation~\cite{shan2024towards,zhao2024dartcontrol,liao2025shape,kim2025personabooth,chen2025language,wu2025text2interact} aims to synthesize human motion that faithfully reflects the intent of a text description.
Beyond producing realistic and fluid motion, a model must capture global semantic structure, such as repetition counts (\emph{``two jumping jacks''}) and sequence-level action patterns (\emph{``turn around, pick things up, and turn around''}).
These semantics govern how the motion unfolds over the entire temporal horizon, not merely local motion dynamics.
Yet, as shown in Fig.~\ref{fig:teaser}, recent diffusion-based models~\cite{zhang2023remodiffuse,zhang2024motiondiffuse}, masked transformers~\cite{pinyoanuntapong2024mmm,guo2024momask,guo2025snapmogen}, and next-token AR models~\cite{guo2022tm2t,zhang2023generating,zhong2023attt2m,zou2024parco} consistently struggle to preserve such global semantics.

We argue that this challenge stems from how existing models construct motion sequences.
Diffusion models and masked transformers begin by generating a full-resolution motion sequence without causal ordering, and then refine it with denoising or masked reconstruction. 
Inferring global semantic structure in this single, undirected prediction step is inherently difficult, especially under the limited size and diversity of current text–motion datasets~\cite{plappert2016kit,guo2022generating}. The subsequent refinement primarily enforces local semantic coherence through bidirectional context, but it does not substantially alter the global structure established in the initial draft. Consequently, the generated motions often remain plausible yet fail to follow the text with the desired level of semantic precision.

AR models have achieved remarkable success in language modeling~\cite{achiam2023gpt,touvron2023llama,openai2025gpt5}, where next-token prediction supports strong long-range reasoning~\cite{rae2019compressive}. In text-to-motion generation, however, next-token AR underperforms and shows a larger gap compared to masked transformers in both motion quality and text alignment.
This gap arises from the characteristics of human motion: its highly predictable short-term dynamics allow each future pose to be inferred from only a brief history~\cite{martinez2017human}. During training, AR models minimize loss by exploiting this short-horizon predictability rather than learning the global temporal structure. Temporal convolution encoders in motion tokenization~\cite{van2017neural} further amplify local correlations, making long-range reasoning even less necessary.
As a result, next-token AR produces locally coherent motion but does not reliably preserve global intent of texts.

To overcome these limitations, we propose MoScale, a next-scale autoregressive framework that replaces next-token prediction with a coarse-to-fine causal hierarchy. Inspired by recent advances in next-scale modeling~\cite{tian2024visual}, MoScale adapts this idea to human motion by introducing a hierarchy aligned with the semantic structure of text-conditioned motion, using only standard T5~\cite{raffel2020exploring} features without any task-specific text engineering. At the coarsest scale, the model commits to the global organization of the sequence, and finer scales progressively refine it at higher temporal resolutions. This hierarchical causality removes the short-horizon shortcut of next-token AR and provides a global semantic scaffold that diffusion- and masked-transformer–based refinements cannot reliably establish.

To further enhance MoScale, we introduce two refinement designs that address limitations of standard next-scale autoregression and the data-scarce nature of text-to-motion generation.
(1) \emph{Cross-scale hierarchical refinement}.
Unlike next-scale autoregression in \cite{tian2024visual}, where each scale predicts a fixed residual target, finer scales in MoScale must correct both the quantization residual and the prediction errors produced at coarser scales. We therefore dynamically adjust the learning target by perturbing coarse-scale predictions during training and supervising finer scales to refine these imperfect drafts. This strategy exposes the model to perturbed intermediate states and enables finer scales to correct errors propagated from earlier stages, reducing exposure bias and improving stability in the hierarchical generation process.
(2) \emph{In-scale temporal refinement}.
Text–motion datasets~\cite{plappert2016kit,guo2022generating} are far smaller than language corpora, and recent studies~\cite{ni2025difflm,prabhudesai2025diffusion} show that diffusion-style iterative refinement and bidirectional context improve robustness in low-data regimes. Motivated by these findings, MoScale revisits predictions within each scale using a selective mask-and-repredict procedure. Bidirectional attention is inherently available within each scale to assess token reliability and selectively re-predict uncertain regions. This improves temporal coherence and local fidelity while maintaining the coarse-to-fine causal structure.

Experiments show that MoScale achieves state-of-the-art performance on standard text-to-motion benchmarks while retaining the high training efficiency characteristic of autoregressive models~\cite{brown2020language,touvron2023llama}. As shown in Fig.~\ref{fig:teaser}, MoScale follows fine-grained text instructions substantially better than prior methods. Our ablation studies reveal that the hierarchical next-scale formulation is the primary driver of improved text alignment. Removing hierarchical refinement yields alignment performance comparable to the recent MoMask++~\cite{guo2024momask}, while restoring it provides a large improvement, whereas temporal refinement contributes mainly to local temporal coherence.
Beyond text-to-motion generation, MoScale supports zero-shot generalization to a range of conditional and unconditional motion tasks, including inpainting, outpainting, editing, and continuous motion generation, while preserving unedited regions. User studies consistently favor MoScale over previous approaches in both text-to-motion and other tasks, highlighting its advantages in both realism and semantic fidelity.

In summary, our contributions are:
\begin{enumerate}
\item We introduce a next-scale AR that overcomes the limitations of next-token AR for motion generation.
\item We introduce cross-scale hierarchical refinement and in-scale temporal refinement to improve motion quality.
\item MoScale achieves state-of-the-art performance, scales with model size, and supports several motion generation tasks in a zero-shot manner.
\end{enumerate}

\section{Related Work}
\textbf{Human Motion Generation.} 
Research on human motion generation, particularly in the text-conditioned setting, has evolved from VAE-based models~\cite{guo2020action2motion,guo2022generating,petrovich2021action,petrovich2022temos}, which learn regularized latent spaces but often struggle with diversity and text alignment, to autoregressive and diffusion-based methods.
Although short-term motion continuity helps produce smooth motion~\cite{fragkiadaki2015recurrent,pavllo2018quaternet}, it also creates a shortcut that lets models favor locally plausible dynamics over long-range semantic alignment.
AR methods~\cite{guo2022tm2t,zhang2023generating,jiang2023motiongpt,zhong2023attt2m,wang2024motiongpt,zheng2025scalable} tokenize motion with VQ-VAE~\cite{van2017neural} and predict tokens sequentially, but can over-rely on short motion history and are further limited by the mismatch between bidirectional token dependencies and unidirectional causal modeling.
Diffusion-based approaches generate motion by iterative denoising in the continuous motion space~\cite{tevet2022human,kim2023flame,zhang2023remodiffuse,zhang2024motiondiffuse,chen2023executing}, alleviating AR causality limitations through bidirectional attention and refinement. However, their performance remains comparable rather than clearly superior. \cite{meng2024rethinking} attributes it to the mismatch between motion data distributions and Gaussian diffusion assumptions.
Masked transformers~\cite{pinyoanuntapong2024mmm,guo2024momask,guo2025snapmogen} have been explored to unify token representations with diffusion-style modeling. MMM~\cite{pinyoanuntapong2024mmm} uses bidirectional attention to iteratively reconstruct masked tokens, avoiding causality mismatch. MoMask~\cite{guo2024momask} adopts residual quantization to generate multi-layer token sequences, but later layers contribute little. MoMask++~\cite{guo2025snapmogen} streamlines with a shared codebook and a unified transformer. However, it applies random token perturbations across layers in training, breaking hierarchical causality.
Like diffusion models, masked transformers remain influenced by motion continuity, which biases learning toward local consistency rather than global text alignment.
In contrast, we show that introducing hierarchical causality into AR modeling enables more effective capture of global motion structure, resulting in improved semantic alignment and generation quality.
Another line of work improves text-conditioned performance through hard word mining~\cite{jeong2025hgm3} and elaborate architectures~\cite{jin2023act,wang2023fg} for more expressive text representations. Instead, we simply adopt T5 features~\cite{raffel2020exploring} with standard attention-based fusion~\cite{lai2022adafusion,ma2024velovox}.
Recent work has also shown the value of modalities beyond text~\cite{lan2024acoustic,lan2025guiding}. Such multimodal conditioning is also being explored in motion generation~\cite{xu2025mospa,yang2025unimumo} and could benefit from our next-scale autoregressive model.
% In contrast, our method follows hierarchical causality through a coarse-to-fine autoregressive framework, achieving better performance while incorporating diffusion-style iterative refinement within each scale.
% Within each scale, we incorporate iterative refinement inspired by diffusion methods to enhance temporal consistency without breaking the structured generation process.

\noindent\textbf{Modeling with Limited Data.}
Autoregressive models have long been the standard in large language modeling, but recent studies~\cite{ni2025difflm,prabhudesai2025diffusion} reveal that diffusion models outperform in low-data scenarios. \cite{ni2025difflm} explains this advantage by factors like increased per-sample computation through iterative refinement and diverse data augmentation. \cite{prabhudesai2025diffusion} highlights the benefits of randomized masking, which trains models over diverse token orderings, unlike the fixed left-to-right order in autoregressive models.
However, motion generation presents different challenges due to its temporal continuity and differences in tokenization.
In this domain, ScaMo~\cite{lu2025scamo} investigates scaling laws for AR models using a large curated dataset, but does not explore generalization in limited-data settings~\cite{plappert2016kit,guo2022generating} and remains constrained to previous next-token prediction.
In this work, we revisit autoregressive modeling through a next-scale framework for motion, integrating diffusion-style benefits within each scale to improve generalization with limited data.
\section{Preliminaries}
\label{sec:preliminaries}
\noindent\textbf{Next Token Prediction.} 
AR models generate sequences by predicting tokens sequentially along the temporal or spatial dimension, where tokens are typically obtained with Vector Quantized VAE (VQ-VAE)~\cite{van2017neural} which represent image patches~\cite{chen2020generative} or motion frames~\cite{zhang2023generating,jiang2023motiongpt}.
Given a quantized discrete token sequence $(x_1, \dots, x_N)$, the model predicts each token based on its prefix and optional conditioning input $\mathbf{c}$, modeling the joint distribution as:
\begin{equation}
    p(x_1, \dots, x_N) = \prod_{n=1}^{N} p(x_n \mid x_1, \dots, x_{n-1}, \textbf{c}).
\end{equation}
This formulation assumes a strict causal ordering. In practice, however, VQ-VAE encoders adopt temporal or spatial convolutions, causing features to depend on both past and future positions. Consequently, the quantized tokens show bidirectional dependencies, creating a mismatch with the unidirectional assumptions of next-token AR models.

\noindent\textbf{Next Scale Prediction.}  
To address this causality mismatch, recent work~\cite{tian2024visual} proposes modeling sequences across hierarchical scales for images. In this setting, the sequence is organized as a hierarchy of token groups $(\mathbf{z}_1, \dots, \mathbf{z}_K)$, where each $\mathbf{z}_k$ captures information at a different resolution. Coarser scales encode global structure, while finer scales represent residual details not captured by earlier levels with a residual VQ-VAE.
Unlike next-token prediction, which generates one token at a time, next-scale prediction generates the entire group $\mathbf{z}_k$ together, conditioned on previously generated coarser groups and conditioning input $\mathbf{c}$:
\begin{equation}
\label{eqn:nextscale}
    p(\mathbf{z}_1, \dots, \mathbf{z}_K) = \prod_{k=1}^{K} p(\mathbf{z}_k \mid \mathbf{z}_1, \dots, \mathbf{z}_{k-1}, \textbf{c}).
\end{equation}
% This coarse-to-fine approach shifts generation from token-level to scale-level dependencies, aligning the residual hierarchy with the unidirectional causality of AR models.

% \noindent\textbf{Encoder}
% \noindent\textbf{Autoregressive}

\section{Method}
\begin{figure*}[t]
    \centering
    \includegraphics[width=1\linewidth]{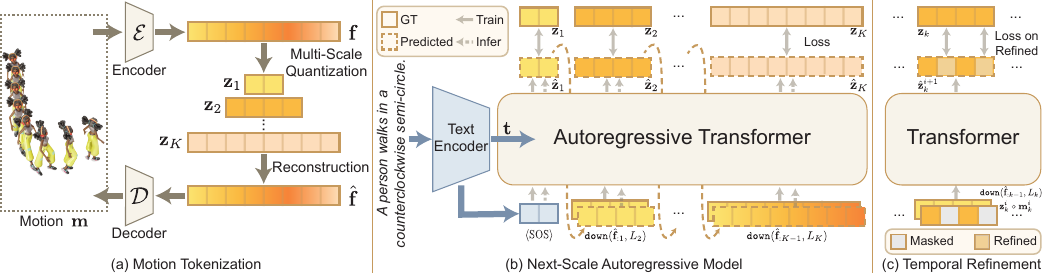}
    \vspace{-20pt}
    \caption{Overview of MoScale. (a) MoScale encodes motion sequences into discrete tokens from coarse to fine through multi-scale quantization. (b) It autoregressively predicts tokens at the next scale, conditioned on the prefix and text inputs, using hierarchical scale-wise causal attention. (c) Within each scale, MoScale performs temporal refinement to further improve token quality and consistency.}
    \vspace{-15pt}
    \label{fig:method}
\end{figure*}
In this section, we first introduce the next-scale prediction framework for motion generation (Sec.~\ref{subsec:next-scale}). We then detail two refinement components: cross-scale hierarchical refinement (Sec.~\ref{subsec:train-inferGAP}) and in-scale temporal refinement (Sec.~\ref{subsec:multi-infer}).

\subsection{Next-Scale Motion Generation}
\label{subsec:next-scale}
\noindent\textbf{Hierarchical Motion Representation.}
Our next-scale motion generation framework builds on a residual VQ-VAE that hierarchically encodes motion into $K$ scales of discrete tokens ~(Fig.~\ref{fig:method}(a)).
Given an input motion sequence $\mathbf{m} \in \mathbb{R}^{T \times D_m}$, where $T$ is the sequence length and $D_m$ is the motion dimension, an encoder $\mathcal{E}$ encodes it into $\mathbf{f} = \mathcal{E}(\mathbf{m}) \in \mathbb{R}^{T/l \times D_e}$, where $l$ is the temporal downsampling rate and $D_e$ the latent feature dimension.

To quantize the latent features $\mathbf{f}$, we first define a set of $K$ increasing lengths $(L_1, L_2, \dots, L_K)$ corresponding to different temporal resolutions, and a shared codebook $\mathbf{Z} \in \mathbb{R}^{V \times D_e}$ of size $V$. At each scale $k$, we first compute a target feature $\mathbf{f}_k \in \mathbb{R}^{L_k \times D_e}$ by downsampling the residual not captured by the previous levels to the current scale:
\begin{equation}
\mathbf{f}_k =
\begin{cases}
\mathtt{down}(\mathbf{f}, L_1), 
    & \text{if } k = 1, \\
\mathtt{down}\!\left(\mathbf{f} - \hat{\mathbf{f}}_{:k-1},\, L_k\right), 
    & \text{otherwise}.
\end{cases}
\end{equation}
where $\hat{\mathbf{f}}_{:k-1} = \sum_{i=1}^{k-1} \hat{\mathbf{f}}_i$ denotes the accumulated reconstruction from all previous $k-1$ scales, $\hat{\mathbf{f}}_i$ denotes the dequantized and upsampled feature from tokens at scale $i$, defined in Eqn.~\ref{eqn:upsample}.
Each $\mathbf{f}_k$ is then quantized by the codebook:
\begin{equation}
\mathbf{z}_k = \arg\min_{v \in [V]} \|\mathbf{f}_k - \mathbf{Z}_v\|_2^2,
\end{equation}
resulting in a discrete token sequence $\mathbf{z}_k \in [V]^{L_k}$. 

To obtain the reconstructed feature $\hat{\mathbf{f}}_k$ from this scale, we map tokens back to embeddings via a $\mathtt{lookup}()$ operation and then upsample them to the original latent length $T/l$:
\begin{equation}
\label{eqn:upsample}
\hat{\mathbf{f}}_k = \mathtt{up}(\mathtt{lookup}(\mathbf{z}_k), T/l).
\end{equation}
To reconstruct the full motion sequence, the decoder $\mathcal{D}$ aggregates the dequantized features from all scales $\hat{\mathbf{f}} = \hat{\mathbf{f}}_{:K}$ and maps them back to the motion space:
$
\hat{\mathbf{m}} = \mathcal{D}\left( \hat{\mathbf{f}} \right).
$

% This tokenization acheves a hierarchy of discrete token groups $(\mathbf{z}_1, \dots, \mathbf{z}_K)$, where each group encodes motion at a distinct temporal resolution. By structuring the generation such that each level conditions only on coarser scales, the model establishes a coarse-to-fine causal hierarchy. 

% Each scale incorporates textual attention to ensure that the aggregated feature $\hat{\mathbf{f}}$ aligns with the input description.

%The model is trained with a combination of reconstruction loss between the reconstructed motion $\hat{\mathbf{m}}$ and the ground truth $\mathbf{m}$, an embedding loss that aligns residual features $\mathbf{f}_k$ with their quantized embeddings $\tilde{\mathbf{z}}_k$, and a velocity regularization term~\cite{zhang2023generating} to encourage temporal smoothness.

\noindent\textbf{Text to Motion Generation.}
Given a motion represented as a hierarchy of discrete token groups $(\mathbf{z}_1, \dots, \mathbf{z}_K)$, we leverage a causal transformer to autoregressively generate these tokens conditioned on an input text description, as shown in Fig.~\ref{fig:method}(b).
Specifically, we first encode texts with a pretrained T5 encoder~\cite{raffel2020exploring}, obtaining text embeddings $\mathbf{t} \in \mathbb{R}^{S \times D_t}$, where $S$ is the number of tokens and $D_t$ is the dimension. The embeddings are then processed via attention-based pooling and projected to $T \in \mathbb{R}^{D}$ , where $D$ denotes the transformer latent dimension. We then broadcast it to $\langle \text{SOS} \rangle \in \mathbb{R}^{L_1 \times D}$ as input to the first scale, where the transformer is required to predict $\mathbf{z}_1$ based on it. At each following scale $k$, we downsample the accumulated feature $\hat{\mathbf{f}}_{:k-1}$ from the previous scales to length $L_k$ to match the resolution of scale $k$. At each scale, the output logits are projected to the codebook size $V$, yielding the probability of each token in the codebook $Z$.

Our transformer consists of $M$ blocks, where every block includes a self-attention layer and a cross-attention layer. 
To ensure hierarchical causality across scales, a scale-wise causal mask is applied in the self-attention, preventing information leakage from higher scales. The cross-attention attends directly to the raw text embeddings $\mathbf{t}$ to maintain consistent semantic guidance throughout the hierarchical generation process.
For positional encoding, we apply normalized rotary position embeddings (RoPE)~\cite{su2024roformer,ma2024star} with a learnable scale-level embedding to the input of each block, facilitating positional relationships across scales.

\subsection{Cross-Scale Hierarchical Refinement}
\label{subsec:train-inferGAP}
While next-scale autoregression aligns naturally with the causal structure of motion and achieves a high training efficiency, it remains limited in its ability to correct errors propagated across scales. We attribute this to a combination of limited training diversity and the use of teacher forcing, which exposes the model only to clean, ground-truth inputs at each level. As a result, the model fails to learn how to recover from imperfect coarser-scale predictions during inference, leading to error accumulation and degraded motion.

To address this limitation, we propose a cross-scale hierarchical refinement that facilitates adaptive residual learning across scales.
During training, for scale $k$, we introduce perturbation to the tokens of the coarser scale $k-1$. Specifically, a uniformly sampled subset of tokens in $\mathbf{z}_{k-1}$ is replaced with random tokens in codebook $Z$, forming a corrupted sequence $\tilde{\mathbf{z}}_{k-1}$. Throughout this paper, we use a tilde (e.g., $\tilde{\mathbf{z}}$) to denote perturbed variables. This sequence is then dequantized and upsampled to obtain $\tilde{\hat{\mathbf{f}}}_{k-1}$, which replaces the clean feature $\hat{\mathbf{f}}_{k-1}$ in the accumulation:
\begin{equation}
\tilde{\hat{\mathbf{f}}}_{:k-1} = \sum_{i=1}^{k-2} \hat{\mathbf{f}}_i + \tilde{\hat{\mathbf{f}}}_{k-1},
\end{equation}
We then define the target residual at scale $k$ as: 
\begin{equation}
\tilde{\mathbf{f}}_k = \mathtt{down}\left(\mathbf{f} - \tilde{\hat{\mathbf{f}}}_{:k-1}, L_k\right),
\end{equation}
and quantize it to obtain training targets $\tilde{\mathbf{z}}_k$ for scale $k$.
Note that the perturbation is applied to the input of the transformer at scale $k$, and does not affect the learning target for scale $k-1$, which remains ${\mathbf{z}}_{k-1}$. This design simulates prediction errors from scale $k-1$. 
However, unlike on-the-fly prediction methods that rely on sequential supervision, our design allows efficient single-pass training with dense multi-scale supervision.
Importantly, this aligns with the residual causal structure where each scale learns to encode the information not captured by earlier ones, and thus perturbing scale $k$ does not affect the supervision of scale $k-1$, which is responsible only for the residuals left by scales $(1, \cdots, k-2)$.
Our refinement strategy is applied across all scales. For each training sample, we randomly corrupt a subset of tokens in every $\mathbf{z}_k$, with a corruption ratio $\gamma_k$ sampled uniformly from $[0, \gamma_{\max}]$.

\noindent\textbf{Discussion.}
This design enables the model to learn adaptive and robust hierarchical refinements under imperfect contexts, especially important in limited-data regimes. By exposing the model to a range of corrupted inputs while maintaining fixed supervision targets, it learns to consistently infer residuals that recover the feature $\textbf{f}$. Additionally, varying the number and combination of active scales during training further improves generalization by encouraging the model to adapt to diverse intermediate contexts.

\subsection{In-Scale Temporal Refinement}
\label{subsec:multi-infer}
To enhance temporal coherence and motion quality, we refine token predictions within each scale with an iterative process. While cross-scale generation operates under a coarse-to-fine causal hierarchy, tokens within each individual scale are bidirectionally dependent. Motivated by recent findings in low-data regimes~\cite{ni2025difflm,prabhudesai2025diffusion}, we leverage the bidirectional attention already present in the next-scale autoregressive model to capture these dependencies. Specifically, we use a selective mask-and-repredict strategy that identifies uncertain tokens and iteratively refines them without disrupting the global structure established at coarser levels.

As illustrated in Fig.~\ref{fig:method}(c), let $\mathbf{z}_k^i$ denote the predicted tokens at refinement iteration $i$. An element-wise binary mask $\mathbf{m}_k^i \in \{0, 1\}^{L_k}$ is constructed, where entries are set to 0 for low-confidence tokens to be re-predicted in the next iteration, while confident tokens are retained. To enable this, we concatenate the accumulated features ${\hat{\mathbf{f}}}_{:k-1}$ from previous scales with the masked token sequence $\mathbf{z}_k^i \circ \mathbf{m}_k^i$, where $\circ$ denotes element-wise multiplication. This combined input is fed into the transformer, which is trained to reconstruct the masked tokens conditioned on the visible context.
In practice, masked tokens are replaced with a special $\texttt{[MASK]}$ token. We adopt a cosine re-masking schedule~\cite{chang2022maskgit} and specify the number of refinement steps per scale as $(I_1, \dots, I_K)$, allowing the model to enhance prediction quality.
This refinement introduces diffusion-style iterative correction into the autoregressive framework, enabling fine-grained updates and improved temporal consistency. 

% Beyond this, the temporal refinement also enables powerful zero-shot capabilities of MoScale for motion editing, inpainting, outpainting, and continuous generation, applicable to both conditional and unconditional settings. These tasks are unified under the same mask-and-repredict framework by simply masking tokens in regions to be modified and keeping the remaining tokens fixed. To minimize changes in uninterested regions, we retain tokens for unedited areas across all scales rather than only at the coarsest level. This enables the model to localize modifications more precisely, resulting in more precise and effective motion edits, as demonstrated in Sec.~\ref{subsec:zeroshot}.
% The model then uses this mask to selectively re-predict only the uncertain tokens at the next iteration, while preserving high-confidence predictions. This iterative process improves local temporal consistency without breaking global causality.

Apart from this, temporal refinement also empowers MoScale with strong zero-shot capabilities for tasks such as motion editing, inpainting, outpainting, and continuous generation, across both conditional and unconditional settings. These tasks are unified under the same mask-and-repredict strategy, where target regions are edited by masking the corresponding tokens while keeping the rest fixed.
To better preserve content in unedited regions, we retain tokens for those areas across all scales, rather than only at the coarsest level. This design enables the model to localize modifications more precisely and apply changes in a more controlled and fine-grained manner, resulting in more effective motion edits, as demonstrated in Sec.~\ref{subsec:zeroshot}.

\subsection{Model Training and Inference}
\noindent\textbf{Training.}
% We train our residual VQ-VAE using a combination of a reconstruction loss between $\hat{\mathbf{m}}$ and $\mathbf{m}$, an embedding loss between residual features $\mathbf{f}_k$ and the quantized embeddings $\tilde{\mathbf{z}}_k$, and a velocity regularization term~\cite{zhang2023generating} for temporal smoothness.
We train our residual VQ-VAE using a combination of a reconstruction loss between $\hat{\mathbf{m}}$ and $\mathbf{m}$, an explicit joint position loss for additional supervision on local body joints, and a commitment loss between the encoder outputs and the assigned entries.
To train our causal autoregressive model, we adopt teacher forcing with cross-entropy loss between predicted $\hat{\mathbf{z}}_k$ and ground-truth tokens $\mathbf{z}_k$.
We also apply classifier-free guidance (CFG) by randomly dropping the text condition with a probability of 10\% during training.

\noindent\textbf{Inference.}
During inference, our model first samples tokens $\hat{\mathbf{z}}_1$ at the coarsest scale. It then autoregressively generates tokens for each subsequent scale, conditioned on the previously sampled tokens $(\hat{\mathbf{z}}_1, \dots, \hat{\mathbf{z}}_{k-1})$ and text. This coarse-to-fine process continues until tokens at all scales are generated. The resulting tokens are decoded by $\mathcal{D}$ to produce the output motion sequence.
\section{Experiments}
\begin{table*}[t]
\centering
\small
\setlength{\tabcolsep}{5pt}

\begin{tabular}{l l c c c c c c c}
\toprule
\textbf{Method} & \textbf{Structure} & \textbf{Top1} $\uparrow$ & \textbf{Top2} $\uparrow$ & \textbf{Top3} $\uparrow$ & \textbf{FID} $\downarrow$ & \textbf{MM-Dist} $\downarrow$ & \textbf{Diversity} $\rightarrow$ & \textbf{MModality} $\uparrow$ \\
\midrule
Real Motion              & -            & \numsup{0.511}{.003} & \numsup{0.703}{.003} & \numsup{0.797}{.002} & \numsup{0.002}{.000} & \numsup{2.974}{.008} & \numsup{9.503}{.065} & - \\
\midrule
TM2T~\cite{guo2022tm2t}            & \multirow{4}{*}{Next-Token AR} & \numsup{0.424}{.003} & \numsup{0.618}{.003} & \numsup{0.729}{.002} & \numsup{1.501}{.017} & \numsup{3.467}{.011} & \numsup{8.589}{.076} & \numsup{2.090}{.083} \\
T2M-GPT~\cite{zhang2023generating}  &                             & \numsup{0.492}{.003} & \numsup{0.679}{.002} & \numsup{0.775}{.002} & \numsup{0.141}{.005} & \numsup{3.121}{.009} & \numsup{9.722}{.082} & \numsup{2.424}{.093} \\
AttT2M~\cite{zhong2023attt2m}          &                             & \numsup{0.499}{.003} & \numsup{0.690}{.002} & \numsup{0.786}{.002} & \numsup{0.112}{.006} & \numsup{3.038}{.007} & \numsup{9.700}{.090} & \numsup{1.831}{.048} \\
ParCo~\cite{zou2024parco}          &                             & \numsup{0.515}{.003} & \numsup{0.706}{.003} & \numsup{0.801}{.002} & \numsup{0.109}{.005} & \numsup{2.927}{.008} & \numsup{9.576}{.088} & \numsup{\underline{2.452}}{.051} \\
\midrule
MDM~\cite{tevet2022human}             & \multirow{5}{*}{Diffusion}  & \numsup{0.418}{.005} & \numsup{0.604}{.005} & \numsup{0.707}{.004} & \numsup{0.489}{.025} & \numsup{3.630}{.023} & \numsup{{9.450}}{.066} & \numsup{1.382}{.060} \\
MLD~\cite{chen2023executing}             &                             & \numsup{0.481}{.003} & \numsup{0.673}{.003} & \numsup{0.772}{.002} & \numsup{0.473}{.013} & \numsup{3.196}{.010} & \numsup{9.724}{.082} & \numsup{1.553}{.042} \\
MotionDiffuse~\cite{zhang2024motiondiffuse}   &                             & \numsup{0.491}{.001} & \numsup{0.681}{.001} & \numsup{0.782}{.001} & \numsup{0.630}{.001} & \numsup{3.113}{.001} & \numsup{9.410}{.049} & \numsup{\textbf{2.870}}{.111} \\
ReMoDiffuse~\cite{zhang2023remodiffuse}     &                             & \numsup{0.510}{.005} & \numsup{0.698}{.006} & \numsup{0.795}{.004} & \numsup{0.103}{.004} & \numsup{2.974}{.016} & \numsup{9.018}{.075} & \numsup{2.413}{.079} \\
DiverseMotion~\cite{lou2023diversemotion}   &                             & \numsup{0.515}{.003} & \numsup{0.706}{.002} & \numsup{0.802}{.002} & \numsup{0.072}{.004} & \numsup{2.941}{.007} & \numsup{9.683}{.102} & \numsup{1.795}{.043} \\
\midrule
MMM~\cite{pinyoanuntapong2024mmm}             & \multirow{3}{*}{MaskedTrans} & \numsup{0.515}{.002} & \numsup{0.708}{.002} & \numsup{0.804}{.002} & \numsup{0.089}{.005} & \numsup{2.926}{.007} & \numsup{9.577}{.050} & \numsup{1.869}{.089} \\
MoMask~\cite{guo2024momask}          &                             & \numsup{0.521}{.002} & \numsup{0.713}{.002} & \numsup{0.807}{.002} & \numsup{\textbf{0.045}}{.002} & \numsup{2.958}{.008} & -                 & \numsup{1.226}{.040} \\
MoMask++~\cite{guo2025snapmogen}        &                             & \numsup{0.528}{.003} & \numsup{0.718}{.003} & \numsup{0.811}{.002} & \numsup{0.072}{.003} & \numsup{2.912}{.008} & -                 & \numsup{1.241}{.040} \\
\midrule
Ours ($S$=4)   & Next-Scale AR & \numsup{\underline{0.535}}{.003} & \numsup{\underline{0.721}}{.003} & \numsup{\underline{0.812}}{.002} & \numsup{0.049}{.002} & \numsup{\underline{2.867}}{.007} & \numsup{\textbf{9.510}}{.089} & \numsup{0.917}{.037} \\
Ours ($S$=18)  & Next-Scale AR & \numsup{\textbf{0.540}}{.002} & \numsup{\textbf{0.727}}{.002} & \numsup{\textbf{0.817}}{.002} & \numsup{\underline{0.046}}{.002} & \numsup{\textbf{2.830}}{.005} & \numsup{\underline{9.525}}{.090} & \numsup{0.873}{.044} \\
\bottomrule
\end{tabular}
\vspace{-5pt}
\caption{Performance on HumanML3D. We report the average result over 20 runs with 95\% confidence interval. \textbf{Bold} for the best and \underline{underline} for the second. $\rightarrow$ indicates that values closer to real motion correspond to better results. $S$ is the total steps across all scales.}
\vspace{-10pt}
\label{tab:HumanML3Dresult}
\end{table*}

% \subsection{Datasets and Metrics}
\noindent\textbf{Datasets.} We evaluate on two widely used text-to-motion benchmarks: HumanML3D~\cite{guo2022generating} and KIT-ML~\cite{plappert2016kit}.
HumanML3D contains 14,616 motion sequences sourced from AMASS~\cite{mahmood2019amass} and HumanAct12~\cite{guo2020action2motion}, paired with a total of 44,970 text descriptions. All motions are resampled to 20 FPS and capped at a maximum length of 10 seconds.
KIT-ML is a smaller dataset comprising 3,911 motions and 6,278 text descriptions. Motions are drawn from the KIT~\cite{mandery2015kit} and CMU~\cite{cmu_mocap} and are sampled at 12.5 FPS.
For both datasets, we follow the standard protocol, splitting the data into 80\% for training, 15\% for validation, and 5\% for testing.

\noindent\textbf{Metrics.} We follow the evaluation protocol in T2M~\cite{guo2022generating}.
\textit{R-Precision} measures text-motion alignment by reporting Top-k accuracy in motion-to-text retrieval.
\textit{Fréchet Inception Distance (FID)} assesses motion quality by comparing the distribution of generated motions to that of real motions.
\textit{Multimodal Distance (MM-Dist)} captures the semantic alignment between generated motions and their corresponding text descriptions.
We also report \textit{Diversity}, defined as the average Euclidean distance between randomly sampled motion pairs, and \textit{Multimodality (MModality)}, a secondary metric~\cite{guo2024momask}, computed as the average variance of distances among motions generated from the same prompt.

% \subsection{Implementation Details}
\noindent\textbf{Implementation Details.}
Our residual VQ-VAE has a downsampling rate of 4, a codebook of size $512 \times 512$, and 4 hierarchical scales of discrete tokens, where the sequence lengths at each scale are set as (6, 12, 24, 49) for both HumanML3D and KIT-ML datasets. Our transformer consists of 16 blocks with a latent dimension of 768 and 8 attention heads. During training, we set the maximum token corruption ratio for cross-scale hierarchical refinement to 0.6. The model is trained for 120 epochs on HumanML3D and 60 epochs on KIT-ML, with learning rates of $3 \times 10^{-4}$ and $2 \times 10^{-4}$ respectively, using a linear warm-up schedule. For in-scale temporal refinement, we use a fixed number of refinement steps per scale, set to (1, 2, 5, 10). Finally, classifier-free guidance is set with scales of 5 for HumanML3D and 3 for KIT-ML dataset.

\subsection{Text-to-Motion Performance}
\label{subsec:comparet2m}
\textbf{Quantitative Results.} We compare MoScale against state-of-the-art text-to-motion methods, including next-token AR approaches~\cite{guo2022tm2t,zhang2023generating,zhong2023attt2m,zou2024parco}, diffusion-based methods~\cite{tevet2022human,chen2023executing,zhang2024motiondiffuse,zhang2023remodiffuse,lou2023diversemotion}, and latest masked transformers~\cite{pinyoanuntapong2024mmm,guo2024momask,guo2025snapmogen}. The results are provided in Table~\ref{tab:HumanML3Dresult} and Table~\ref{tab:KIT-MLResult}.
Our next-scale AR method outperforms previous baselines, achieving the best R-Precision and MM-Dist on both HumanML3D and KIT-ML.
It also obtains the best Diversity on HumanML3D and comparable FID on both datasets, achieving new SOTA performance with next-scale AR models.
% \begin{figure}[t]
%     \centering
%     \includegraphics[width=1\linewidth]{figure/easy.pdf}
%     \vspace{-20pt}
%     \caption{{Low, medium and high of text complexity} Tables increase compard with the. \textcolor{red}{use cdf format, change to table, together with the vlm evaluated result}% CDF for 10; 30; 60; 100; Robin
%     }
%     % \caption{Text alignment comparison of differnet methods on samples of different level. Our method achievs a larger performance gap when in high detials requried text to motion instanes. {Low, medium and high of text complexity} Tables increase compard with the. \textcolor{red}{use cdf format, change to table, together with the vlm evaluated result}% CDF for 10; 30; 60; 100; Robin
%     % }
%     \label{fig:trainperform}
%     \vspace{-15pt}
% \end{figure}

\noindent\textbf{Qualitative Results.} We compare with open-sourced SOTA models from three categories: ParCo~\cite{zou2024parco} (next-token AR),  ReMoDiffuse~\cite{zhang2023remodiffuse} (diffusion), and MoMask++~\cite{guo2025snapmogen} (masked transformer). As shown in Fig.~\ref{fig:teaser}, MoScale more faithfully captures fine-grained semantics, e.g., \textit{``two jumping jacks''} and \textit{``turn around, pick things up, and turn around''}, where the baseline methods failed. We provide additional visualizations on the webpage.

\noindent\textbf{User Study on Text Alignment and Motion Quality.}
To complement the quantitative results with human perception, we conduct a user study comparing MoScale with the three baselines as well. We generate 20 motion sequences using text descriptions from HumanML3D, and 20 participants evaluate each sequence on text alignment and motion quality by choosing the one they think performs best.
MoScale is most frequently preferred for text alignment, with 71.5\% of votes, and also receives the highest preference for motion quality, selected as the best by 73.3\% of users.

\begin{table*}[t]
\centering
\small
\setlength{\tabcolsep}{5pt}
\begin{tabular}{l l c c c c c c c}
\toprule
\textbf{Method} & \textbf{Structure} & \textbf{Top1} $\uparrow$ & \textbf{Top2} $\uparrow$ & \textbf{Top3} $\uparrow$ & \textbf{FID} $\downarrow$ & \textbf{MM-Dist} $\downarrow$ & \textbf{Diversity} $\rightarrow$ & \textbf{MModality} $\uparrow$ \\
\midrule
Real Motion              & -               & \numsup{0.424}{.005} & \numsup{0.649}{.006} & \numsup{0.779}{.006} & \numsup{0.031}{.004} & \numsup{2.788}{.012} & \numsup{11.080}{.097} & - \\
\midrule
TM2T~\cite{guo2022tm2t}            & \multirow{4}{*}{Next-token AR} & \numsup{0.280}{.005} & \numsup{0.463}{.006} & \numsup{0.587}{.005} & \numsup{3.599}{.153} & \numsup{4.591}{.026} & \numsup{9.473}{.117}  & \numsup{\textbf{3.292}}{.081} \\
T2M-GPT~\cite{zhang2023generating}  &                              & \numsup{0.416}{.006} & \numsup{0.627}{.006} & \numsup{0.745}{.006} & \numsup{0.514}{.029} & \numsup{3.007}{.023} & \numsup{10.921}{.108} & \numsup{1.570}{.039} \\
AttT2M~\cite{zhong2023attt2m}          &                              & \numsup{0.413}{.006} & \numsup{0.632}{.006} & \numsup{0.751}{.006} & \numsup{0.870}{.039} & \numsup{3.039}{.021} & \numsup{10.960}{.123} & \numsup{\underline{2.281}}{.047} \\
ParCo~\cite{zou2024parco}          &                              & \numsup{0.430}{.004} & \numsup{0.649}{.007} & \numsup{0.772}{.006} & \numsup{0.453}{.027} & \numsup{2.820}{.028} & \numsup{10.950}{.094} & \numsup{1.245}{.022} \\
\midrule
MDM~\cite{tevet2022human}             & \multirow{5}{*}{Diffusion}   & -                     & -                     & \numsup{0.396}{.004} & \numsup{0.497}{.021} & \numsup{9.191}{.022} & \numsup{10.847}{.109} & \numsup{1.907}{.214} \\
MLD~\cite{chen2023executing}             &                              & \numsup{0.390}{.008} & \numsup{0.609}{.008} & \numsup{0.734}{.007} & \numsup{0.404}{.027} & \numsup{3.204}{.027} & \numsup{10.800}{.117} & \numsup{2.192}{.071} \\
MotionDiffuse~\cite{zhang2024motiondiffuse}   &                              & \numsup{0.417}{.004} & \numsup{0.621}{.004} & \numsup{0.739}{.004} & \numsup{1.954}{.062} & \numsup{2.958}{.005} & \numsup{\textbf{11.100}}{.143}  & \numsup{0.730}{.013} \\
ReMoDiffuse~\cite{zhang2023remodiffuse}     &                              & \numsup{0.427}{.014} & \numsup{0.641}{.004} & \numsup{0.765}{.055} & \numsup{\textbf{0.155}}{.006} & \numsup{2.814}{.012} & \numsup{10.800}{.105}  & \numsup{1.239}{.028} \\
DiverseMotion~\cite{lou2023diversemotion}   &                              & \numsup{0.416}{.005} & \numsup{0.637}{.008} & \numsup{0.760}{.011} & \numsup{0.468}{.098} & \numsup{2.892}{.041} & \numsup{10.873}{.101} & \numsup{2.062}{.079} \\
\midrule
MMM~\cite{pinyoanuntapong2024mmm}             & \multirow{2}{*}{MaskedTrans} & \numsup{0.404}{.005} & \numsup{0.621}{.005} & \numsup{0.744}{.004} & \numsup{0.316}{.028} & \numsup{2.977}{.019} & \numsup{10.910}{.101} & \numsup{1.232}{.039} \\
MoMask~\cite{guo2024momask}          &                              & \numsup{0.433}{.007} & \numsup{0.656}{.005} & \numsup{0.781}{.005} & \numsup{0.204}{.011} & \numsup{2.779}{.022} & -                     & \numsup{1.131}{.043} \\
\midrule
Ours ($S$=4)  & Next-Scale AR     & \numsup{\underline{0.441}}{.005} & \numsup{\underline{0.662}}{.006} & \numsup{\underline{0.783}}{.006} & \numsup{0.183}{.011} & \numsup{\underline{2.758}}{.019} & \numsup{10.812}{.094} & \numsup{0.943}{.042} \\
Ours ($S$=18)  & Next-Scale AR     & \numsup{\textbf{0.442}}{.005} & \numsup{\textbf{0.671}}{.007} & \numsup{\textbf{0.791}}{.008} & \numsup{\underline{0.173}}{.010} & \numsup{\textbf{2.717}}{.017} & \numsup{\underline{10.972}}{.088} & \numsup{0.944}{.048} \\
\bottomrule
\end{tabular}
\vspace{-8pt}
\caption{Performance on KIT-ML test set. \textbf{Bold} for the best result and \underline{underline} for the second best.}
\vspace{-15pt}
\label{tab:KIT-MLResult}
\end{table*}

\begin{table}[t]
\centering
\small
\setlength{\tabcolsep}{5pt}
\newcommand{\numsdown}[2]{\ensuremath{#1_{\scriptsize #2}}}

% ---------- main comparison ----------
\begin{tabular}{lccc}
\toprule
% & \multicolumn{4}{c}{Text Alignment$\uparrow$} \\
% \cmidrule(lr){2-5}
Method & \textsc{FULL} & \textsc{MEDIUM+HIGH} & \textsc{HIGH}  \\
\midrule
ParCo    & 0.801 & 0.778 & 0.709\\
MoMask++ & \numsdown{0.811}{.010\uparrow} & \numsdown{0.802}{.024\uparrow} &
           \numsdown{0.762}{.053\uparrow} \\
Ours  & \numsdown{\textbf{0.817}}{.016\uparrow} &
           \numsdown{\textbf{0.812}}{.034\uparrow} &
           \numsdown{\textbf{0.775}}{.066\uparrow} \\
\bottomrule
\end{tabular}
\vspace{-8pt}
\caption{Quantitative comparison on the HumanML3D test set using the Top-3 accuracy. Test samples are ranked by text complexity, and each method is evaluated on progressively more challenging subsets. Subscripts denote absolute improvements over ParCo.}
\vspace{-18pt}
\label{tab:alignment_detail}
\end{table}

\noindent\textbf{Text Alignment Across Varying Complexity.}
To better evaluate how models handle fine-grained semantic instructions, we categorize each text description in the HumanML3D test set into one of three complexity levels: \textsc{low}, \textsc{medium}, and \textsc{high}. A few-shot prompted LLM assigns these labels by considering both the length of each description and the number of distinct motion details it contains. The resulting distribution follows an approximate 4:2:1 ratio from \textsc{low} to \textsc{high} complexity.
We then evaluate each method on progressively more challenging subsets (the \textsc{full} set, \textsc{medium+high} and \textsc{high}). As shown in Table~\ref{tab:alignment_detail}, performance decreases for all methods as text complexity increases. However, MoScale achieves the largest relative gains over baselines in the high-complexity region, highlighting the effectiveness of its hierarchical next-scale design in capturing complex semantic instructions.
% \noindent\textbf{Text Alignment Across Different Complexity.}
% To evaluate how well our method MoScale aligns motion with text, we divide the HumanML3D test set into three subsets based on text complexity: low, medium, and high. \textcolor{red}{how to perform rank and get top 10\%?}
% We achieve this categorization with a few-shot prompted LLM, detailed in Supplementary Material, which considers not just the length of each description but the number of fine-grained motion details it contains. The resulting split follows an approximate 4:2:1 ratio. After ranking the entire test set from highest to lowest, we evaluate each method on progressively larger portions of this ordered list (Top 10\%, 30\%, 60\%, and 100\%), as shown in Table~\ref{tab:alignment_detail}.
% As the evaluation focuses more heavily on descriptions with higher complexity, the performance of all methods decreases. However, MoScale exhibits consistently larger improvements over the baselines in these challenging subsets, demonstrating stronger alignment with detailed and complex motion descriptions.

\noindent\textbf{Factors Behind Alignment Gains.}
To identify which component contributes most to improved text–motion alignment on details, we employ a vision-language model (VLM)~\cite{openai2025gpt5,wang2025physctrl} to score how well the generated motion matches fine-grained text, extracted from original motion descriptions using an LLM. The VLM assigns a discrete score from 1 (low) to 3 (high), and we report the average over 100 randomly sampled descriptions from the high-complexity subset. The full MoScale model achieves the highest score of 2.14, followed by the variant with only hierarchical refinement (2.09). In contrast, using only temporal refinement yields 1.92, and removing both refinements produces 1.89, comparable to the MoMask++ at 1.90.
While it is hard to explicitly isolate the entire hierarchical next-scale framework, several observations collectively point to the hierarchical prediction path as the dominant source of improved text alignment.
First, the strong performance of both the full model and the hierarchical refinement-only variant indicates that strengthening the hierarchical prediction path is the main contributor to improved text alignment.
Second, even the no-refinement variant, which is trained only with fixed ground-truth instances at each scale, nearly matches bidirectional attention-based MoMask++, despite MoMask++ leveraging additional data augmentation. 
This suggests that the coarse-to-fine causal structure provides a more effective semantic scaffold for representing detailed textual cues than bidirectional iterative refinement alone. Temporal refinement, while helpful for local smoothness, contributes relatively little to the text alignment.

% \begin{figure}[t]
%     \centering
%     \includegraphics[width=1\linewidth]{figure/fid_epoch_comparison.pdf}
%     \vspace{-20pt}
%     \caption{Comparison between T2M-GPT and our method on FID across training epochs. We also provide ablations of our model without both Hierarchical Refinement (HR) and Temporal Refinement (TR), and without TR only. \textcolor{red}{replot}}
%     \label{fig:resfid-mmdist-epoch}
%     \vspace{-5pt}
% \end{figure}
\begin{figure}[t]
    \centering
    \includegraphics[width=0.95\linewidth]{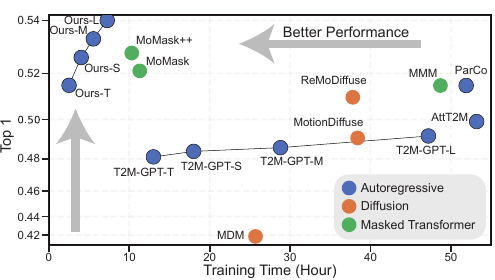}
    \vspace{-8pt}
    \caption{Comparison of Top-1 text alignment and training time on HumanML3D~\cite{guo2022generating}. 
    We provide four model sizes, tiny, small, medium, and large, for our method and T2M-GPT~\cite{zhang2023generating}.}
    %,with each pair matched to a similar parameter scale. Upper left indicates better performance.}
    \label{fig:trainperform}
     \vspace{-18pt}
\end{figure}

\noindent\textbf{Training and Inference Time.} MoScale achieves more efficient training compared to other methods as shown in Fig.~\ref{fig:trainperform}. Unlike AttT2M and ParCo, which extend T2M-GPT with complex modules that increase training cost but offer limited gains, MoScale introduces hierarchical causality and cross-scale refinement, resulting in faster convergence and improved performance as detailed in Sec.~\ref{subsec:AblationStudy}. 
At inference, MoScale takes 0.28s with $S=18$ and 0.08s with $S=4$ (one step per scale) using KV cache~\cite{kwon2023efficient}. Although the full model is slower than MoMask (0.12s), it remains faster than ParCo (1.77s) and substantially more efficient than the diffusion-based MotionDiffuse (7.12s).
The added cost stems from temporal refinement and the longer modeling sequence.
Nevertheless, MoScale significantly improves motion quality and text alignment, and remains practical for typical text-to-motion applications.

% Requires: \usepackage{booktabs,multirow}
% Requires: \usepackage{booktabs}

\subsection{Zero-shot Generalization}
% \begin{figure*}[t]
%     \centering
%     \includegraphics[width=1\linewidth]{figure/EditQual_2.pdf}
%     \vspace{-20pt}
%     \caption{Comparison of motion editing. MoScale better follows the instruction text while preserving the original motion than baselines.}
%     \label{fig:qualedit}
% \end{figure*}
% \label{subsec:zeroshot}

\begin{figure*}[t]
\centering
\begin{tikzpicture}
\node[anchor=south west, inner sep=0.9] (editqual) at (0,0)
{\includegraphics[width=1\linewidth]{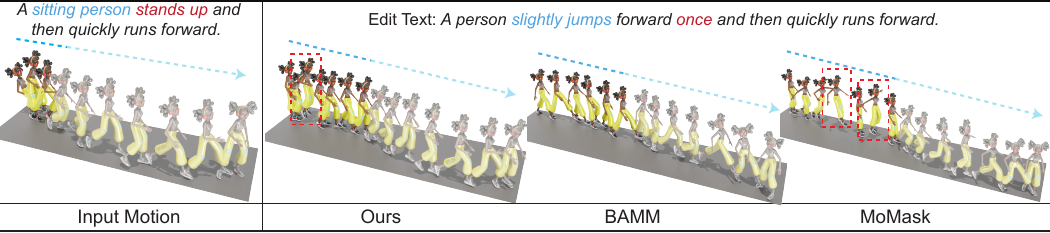}};
\begin{scope}[x={(editqual.south east)}, y={(editqual.north west)}]
    \node[anchor=west, text=black, font=\small, inner sep=0pt, outer sep=0pt]
        at (0.63, 0.075){~\cite{pinyoanuntapong2024bamm}};
    \node[anchor=west, text=black, font=\small, inner sep=0pt, outer sep=0pt]
        at (0.885, 0.075) {~\cite{guo2024momask}};
\end{scope}
\end{tikzpicture}
    \vspace{-20pt}
    \caption{Motion editing results. 
    MoScale achieves better instruction adherence and retains unedited motion (shown in gray).}
    \vspace{-15pt}
\label{fig:qualedit}
\end{figure*}
\label{subsec:zeroshot}

\noindent\textbf{Qualitative Results.}
MoScale supports zero-shot generalization across a range of motion generation tasks. Fig.~\ref{fig:qualedit} showcases an example of motion editing conditioned on text prompts. Our model accurately modifies the motion sequence following the specified instruction, while successfully preserving the unedited regions. Compared to previous methods MoMask~\cite{guo2024momask} and BAMM~\cite{pinyoanuntapong2024bamm}, MoScale produces more faithful and coherent edits with minimal disruption to the surrounding motion. We include additional examples on the webpage for motion editing, inpainting, outpainting, and continuous motion generation. 
% These examples show the effectiveness and flexibility of our framework in practical zero-shot scenarios.

\noindent\textbf{User Study.}
To further evaluate its capability, we conduct a user study on 25 zero-shot cases, covering text-conditioned motion inpainting, outpainting, and editing. Each sample is rated by 10 users across three criteria: motion quality, text alignment, and motion consistency, measuring how well the unedited motion remains intact. Compared to the two baselines BAMM~\cite{pinyoanuntapong2024bamm} and MoMask~\cite{guo2024momask}, MoScale receives the highest preference scores across all metrics, with 78.4\%, 82.0\%, and 78.0\% of votes respectively.

\subsection{Ablation Studies}
\label{subsec:AblationStudy}

\noindent\textbf{Hierarchical and Temporal Refinement.}
Tab.~\ref{tab:ablation_comp} shows the contribution of the two refinement components. Adding hierarchical refinement leads to a substantial performance gain over the baseline, showing its effectiveness in correcting coarse-to-fine residual errors, particularly under limited data availability. In contrast, temporal refinement offers a marginal additional benefit. While combining both mechanisms produces the highest overall scores, the performance boost is largely driven by hierarchical refinement, with temporal refinement playing a secondary role.

\noindent\textbf{Effect of Generation Order.}
To examine the role of generation order, we compare MoScale’s ordered coarse-to-fine generation with an undirected variant built on the same framework, which masks and re-predicts low-confidence tokens over the full sequence across four scales, with the same number of iterations ($18$). The resulting variant achieves a Top-1 score of $\numsup{0.507}{.003}$ and an FID of $\numsup{0.122}{.004}$, and underperforms MoScale. This suggests our ordered coarse-to-fine generation is more effective than undirected generation.

% We compare our coarse-to-fine, ordered MoScale with a variant that masks and repredicts low confidence tokens in an undirected way. We set the iteration count to 18, matching MoScale, and made necessary changes only on MoScale, which achieves Top-1 of $\numsup{0.481}{.003}$ and FID of $\numsup{0.122}{.004}$. The result confirms that coarse-to-fine, ordered motion generation outperforms undirected mask-and-repredict methods.We also report the performance 
% \textcolor{red}{Tab.~\ref{tab:ablation_comp} shows the contribution of the two refinement mechanisms. Hierarchical causality alone provides a strong baseline, but its performance is limited under scarce text–motion data, as it cannot reliably correct errors propagated from coarser scales. Introducing cross-scale hierarchical refinement yields a substantial improvement by enabling adaptive residual correction, already surpassing most prior methods. In-scale temporal refinement further enhances motion quality through selective bidirectional re-prediction. Combining both refinements achieves the best overall performance.}

\noindent\textbf{Corruption Rate.}
We vary the maximum corruption rate $\gamma_{\text{max}} \in \{0.2, 0.4, 0.6, 0.8\}$ for hierarchical refinement. Both motion quality (FID) and text alignment (MM-Dist) improve initially and then degrade as $\gamma_{\text{max}}$ increases, indicating that moderate perturbation provides the best balance between exploration and stability during training.

\noindent\textbf{Transformer Depth.}
We assess model depth using 4, 8, 12, and 16 transformer layers. Performance improves consistently with increasing depth, reflected by lower FID and MM-Dist scores. Notably, even the 4-layer variant achieves strong alignment, highlighting the efficiency of MoScale.

\noindent\textbf{Refinement Iterations and CFG Scale.}
We study refinement iterations by assigning more iterations to finer scales. As shown in Tab.~\ref{tab:iter_only}, performance improves from $(1,1,1,1)$ at first, but further increasing the budget brings no additional gain, showing diminishing returns. Thus, we use $(1,2,5,10)$ by default. For CFG, Fig.~\ref{fig:cfg_scale} shows that Top-1 improves at small guidance scales and then declines, while FID decreases at moderate scales before rising again. We then set the CFG scale to $5$ for the best overall performance.

\begin{table}[t]
\centering
\scriptsize
\setlength{\tabcolsep}{8pt}
\renewcommand{\arraystretch}{1.05}

\begin{tabular}{l c c c}
\toprule
\textbf{Ablation} & \textbf{Top-1} $\uparrow$ & \textbf{FID} $\downarrow$ & \textbf{MM-Dist} $\downarrow$ \\
\midrule
Base      & \numsup{0.481}{.003} & \numsup{0.176}{.007} & \numsup{3.136}{.009} \\
HR        & \numsup{0.534}{.002} & \numsup{0.090}{.005} & \numsup{2.853}{.006} \\
TR        & \numsup{0.497}{.003} & \numsup{0.129}{.005} & \numsup{3.043}{.008} \\
HR \& TR & \numsup{0.540}{.002} & \numsup{0.046}{.002} & \numsup{2.830}{.005} \\

\midrule
$\gamma_{\max}$ = 0.2 & \numsup{0.529}{.003} & \numsup{0.096}{.004} & \numsup{2.858}{.008} \\
$\gamma_{\max}$ = 0.4 & \numsup{0.535}{.002} & \numsup{0.075}{.004} & \numsup{2.837}{.005} \\
$\gamma_{\max}$ = 0.6 & \numsup{0.540}{.002} & \numsup{0.046}{.002} & \numsup{2.830}{.005}  \\
$\gamma_{\max}$ = 0.8 & \numsup{0.534}{.003} & \numsup{0.074}{.003} & \numsup{2.862}{.009} \\

\midrule
4 Layers  & \numsup{0.516}{.003} & \numsup{0.096}{.004} & \numsup{2.955}{.009} \\
8 Layers  & \numsup{0.526}{.002} & \numsup{0.083}{.004} & \numsup{2.909}{.010} \\
12 Layers & \numsup{0.533}{.003} & \numsup{0.071}{.003} & \numsup{2.879}{.009} \\
16 Layers & \numsup{0.540}{.002} & \numsup{0.046}{.002} & \numsup{2.830}{.005}  \\
\bottomrule
\end{tabular}
\vspace{-8pt}
\caption{
Ablation studies. 1) hierarchical and temporal refinement (HR and TR), 2) corruption rate, and 3) transformer depth.}
\vspace{-5pt}
\label{tab:ablation_comp}
\end{table}

% \begin{table}[t]
% \centering
% \scriptsize
% \setlength{\tabcolsep}{4pt}
% \renewcommand{\arraystretch}{1.05}
%  % mean^{std} style
% \begin{tabular}{cc cc cc}
% \toprule
% Iteration \# & \textbf{Top-1} $\uparrow$ & Scale & \textbf{FID} & Scale & \textbf{FID} \\
% \midrule
% (1, 1, 1, 1)  & \numsup{0.535}{.003} & 1  & \numsup{0.146}{.003} & 6  & \numsup{0.054}{.002} \\
% (1, 1, 2, 4)  & \numsup{0.537}{.002} & 2  & \numsup{0.089}{.006} & 7  & \numsup{0.058}{.002} \\
% (1, 2, 3, 6)  & \numsup{0.538}{.002} & 3  & \numsup{0.065}{.003} & 8  & \numsup{0.065}{.004} \\
% (1, 2, 5, 10) & \numsup{0.540}{.002} & 4  & \numsup{0.054}{.003} & 9  & \numsup{0.071}{.004} \\
% (2, 4, 8, 16) & \numsup{0.539}{.002} & \textbf{5}  & \numsup{0.051}{.002} & 10 & \numsup{0.084}{.004} \\
% \bottomrule
% \end{tabular}
% \vspace{-8pt}
% \caption{Ablation studies on refinement iterations and CFG scales.}
% \vspace{-18pt}
% \label{tab:fid_sample}
% \end{table}

\begin{figure}[t]
\centering
\scriptsize
\setlength{\tabcolsep}{6pt}
\renewcommand{\arraystretch}{1.05}
\providecommand{\numsup}[2]{\ensuremath{#1^{\scriptsize #2}}}
\setlength{\abovecaptionskip}{3pt}
\setlength{\belowcaptionskip}{0pt}

\begin{minipage}[t]{0.38\linewidth}
\vspace{0pt}
\centering
\begin{tabular}{lc}
\toprule
Iteration \# & \textbf{Top-1} $\uparrow$ \\
\midrule
(1, 1, 1, 1)  & \numsup{0.535}{.003} \\
(1, 1, 2, 4)  & \numsup{0.537}{.002} \\
(1, 2, 3, 6)  & \numsup{0.538}{.002} \\
(1, 2, 5, 10) & \numsup{0.540}{.002} \\
(2, 4, 8, 16) & \numsup{0.539}{.002} \\
\bottomrule
\end{tabular}
\captionof{table}{Iteration study.}
\label{tab:iter_only}
\end{minipage}
\hspace{0.01\linewidth}
\begin{minipage}[t]{0.56\linewidth}
\vspace{-2pt}
\centering
\includegraphics[width=\linewidth]{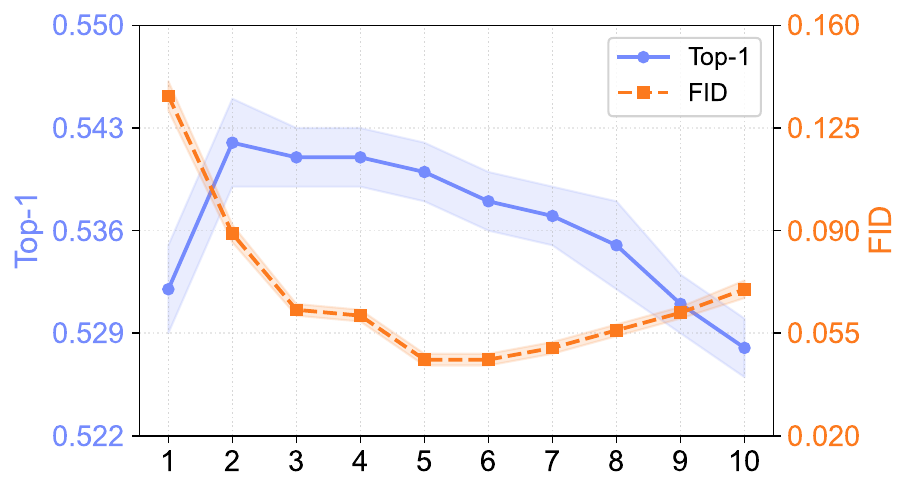}
\vspace{-14pt}
\captionof{figure}{CFG scale.}
\label{fig:cfg_scale}
\end{minipage}
\vspace{-20pt}
\end{figure}
\section{Conclusion}
We present MoScale, an autoregressive framework for hierarchical motion generation that integrates cross-scale hierarchical refinement and in-scale temporal refinement. This design enables high-quality, fine-grained motion synthesis with strong text alignment. Experiments demonstrate that MoScale achieves new SOTA performance and generalizes zero-shot to various motion generation and editing tasks.

% supplementary material
{
    \small
    \bibliographystyle{ieeenat_fullname}
    \bibliography{main}
}

\clearpage
\appendix
\setcounter{page}{1}
\maketitlesupplementary

\begin{strip}
    \centering
    \includegraphics[width=0.83\linewidth]{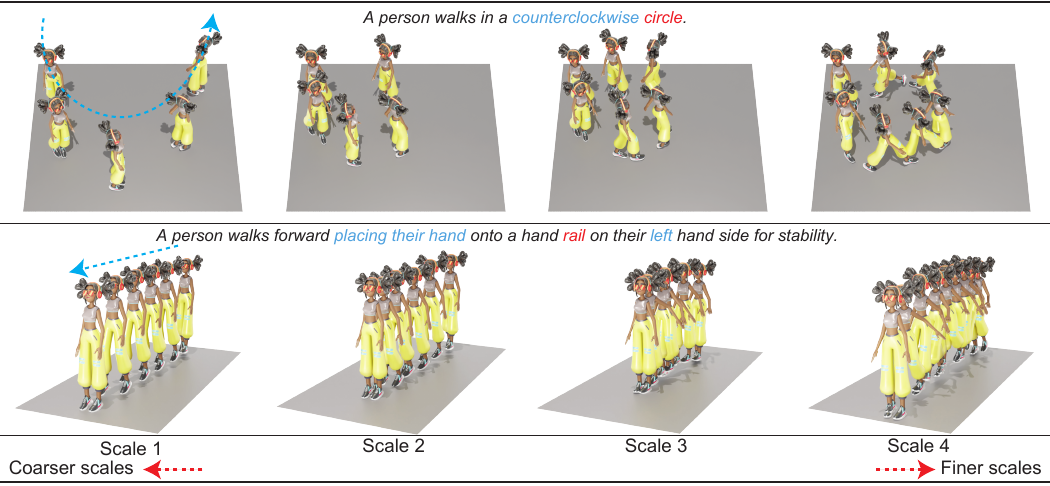}
    \captionof{figure}{Visualization of coarse-to-fine representation of motion with our Residual VQVAE.}
    \label{fig:coarse2fine}
\end{strip}

\section{Coarse to Fine Visualization}
Fig.~\ref{fig:coarse2fine} shows progressive generation from Scale 1 to Scale 4. Scale 1 captures the coarse trajectory and pose but misses fine limb details and exhibits artifacts like sliding. Adding higher scales gradually refines dynamics and articulation, improving local consistency like body orientation and arm placement.
It shows that lower-scale tokens encode coarser structure and that finer-scale tokens provide systematic refinement, illustrating the coarse-to-fine modeling process.

\section{Details of Residual VQVAE}
\subsection{Model Architecture}
\textbf{Encoder.} The encoder maps a raw motion sequence to a compact latent representation. It begins with a linear projection layer that lifts the per-frame features into a higher-dimensional channel space. Two successive downsampling stages then reduce the temporal resolution: each stage applies a convolution followed by a ResNet block and a self-attention block. The attention block uses pre-norm multi-head self-attention with Rotary Position Embeddings (RoPE) and query-dependent gating, followed by a feed-forward layer. Together, the two stages yield an overall 4 times temporal downsampling. A final projection produces the latent sequence.                                            

\noindent\textbf{Hierarchical Residual Quantizer.} We employ a single shared codebook with $K=512$ entries, updated via Exponential Moving Average (EMA). Quantization proceeds over four hierarchical scales, processing the residual signal coarse-to-fine. At each scale, the current residual is temporally downsampled, quantized against the codebook, then upsampled back to the full latent length and accumulated into the reconstructed latent. The final quantized latent of one motion is the sum of all scale contributions. Each motion sequence is thus represented by four index maps at progressively finer temporal resolutions.

\noindent\textbf{Decoder.} The decoder begins with a projection layer, then applies two upsampling stages, each consisting of a dilated ResNet block, upsampling, convolution, and a self-attention block, before a final projection that reconstructs the per-frame motion features.

\noindent\textbf{Loss.} 
We train our residual VQ-VAE using a combination of a reconstruction loss between the original and the reconstructed motion, an explicit joint position loss for additional supervision on local body joints, and a commitment loss between the encoder outputs and the assigned codebook entries. Motion sequences are padded to a fixed maximum length for batched training and inference.

\noindent\textbf{Performance.} 
Our tokenizer achieves strong reconstruction fidelity, with FID of \numsup{0.037}{.001} and MPJPE of \numsup{39.588}{.035} on HumanML3D, and FID of \numsup{0.120}{.003} and MPJPE of \numsup{38.515}{.140} on KIT-ML. While the reconstruction quality is slightly below MoMask~\cite{guo2024momask}, our full model achieves better text-to-motion generation with a smaller gap between reconstruction and generation performance, suggesting that our next-scale autoregressive model preserves and exploits the learned tokens more effectively.

% Our tokenizer achieves strong fidelity, with FID/MPJPE of \numsup{0.037}{.001}/\numsup{39.588}{.035} on HumanML3D and \numsup{0.120}{.003}/\numsup{38.515}{.140} on KIT-ML. While slightly worse than that in MoMask, MoScale yields better text-to-motion generation, with a smaller gap between reconstruction and generation, showing our next-scale AR model preserves and exploits tokens more effectively.

\end{document}